\documentclass[10pt,twocolumn,letterpaper]{article}

\usepackage{iccv}
\usepackage{times}
\usepackage{epsfig}
\usepackage{graphicx}
\usepackage{amsmath}
\usepackage{amssymb}
\usepackage{booktabs}
\usepackage{multirow}
\usepackage{authblk,url}
\renewcommand\footnotemark{}

\usepackage{makecell}

\usepackage{caption}
\usepackage[ruled,linesnumbered]{algorithm2e}

\makeatother

\usepackage[pagebackref=true,breaklinks=true,letterpaper=true,colorlinks,bookmarks=false]{hyperref}

\iccvfinalcopy 

\ificcvfinal\fi
\begin{document}

\title{Convolutional Character Networks}


 \author{Linjie Xing$^{1,2}$, Zhi Tian$^{3}$, Weilin Huang$^{* 1,2}$\thanks{$*$ Corresponding author: whuang@malong.com.}, and Matthew R. Scott$^{1,2}$\\
$^{1}$Malong Technologies, Shenzhen, China \\
$^{2}$Shenzhen Malong Artificial Intelligence Research Center, Shenzhen, China\\
$^{3}$University of Adelaide, Australia\\
}
\maketitle

\begin{abstract}
Recent progress has been made on developing a unified framework for joint text detection and recognition in natural images, but existing joint models were mostly built on two-stage framework by involving ROI pooling, which can degrade the performance on recognition task.
In this work, we propose convolutional character networks, referred as CharNet, which is an one-stage model that can process two tasks simultaneously in one pass. CharNet directly outputs bounding boxes of words and characters, with corresponding character labels. We utilize character as basic element, allowing us to overcome the main difficulty of existing approaches that attempted to optimize text detection jointly with a RNN-based recognition branch. In addition, we develop an iterative character detection approach able to transform the ability of character detection learned from synthetic data to real-world images.  These technical improvements result in a simple, compact, yet powerful one-stage model that works reliably on multi-orientation and curved text. We evaluate CharNet on three standard benchmarks, where it consistently outperforms the state-of-the-art approaches \cite{lyu2018mask,liu2018fots} by a large margin, e.g., with improvements of 65.33\%$\rightarrow$71.08\% (with generic lexicon) on ICDAR 2015, and 54.0\%$\rightarrow$69.23\% on Total-Text, on end-to-end text recognition.
Code is available at: \url{https://github.com/MalongTech/research-charnet}.
\end{abstract}

\section{Introduction}

\begin{figure}[t]
    \begin{center}
        \includegraphics[width=0.98\linewidth]{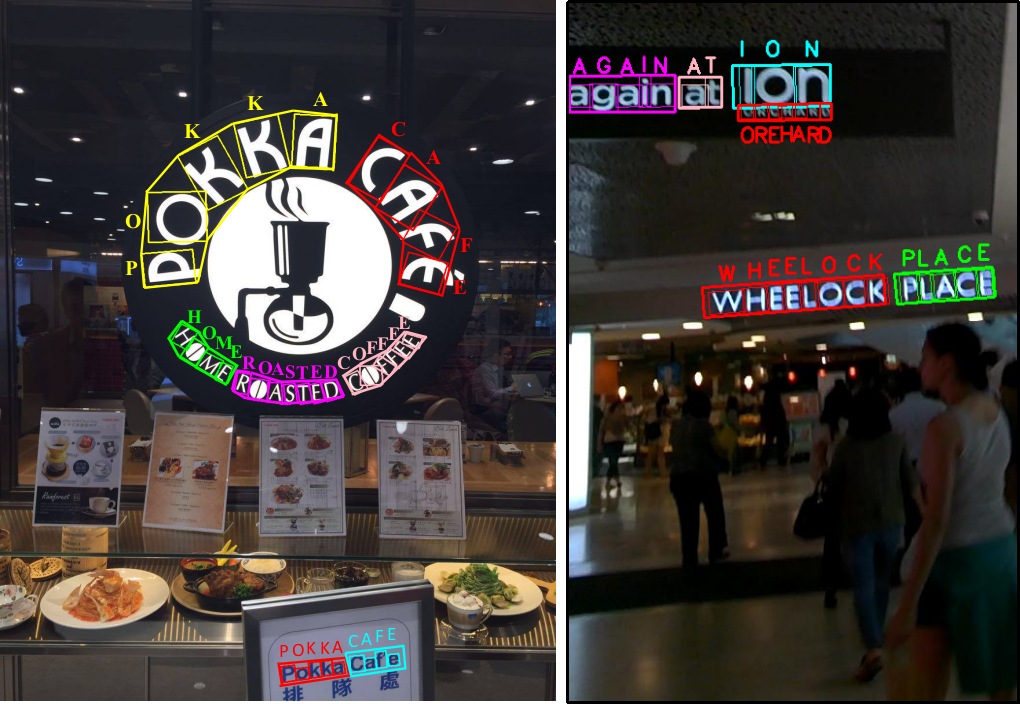}
    \end{center}
    \vspace{-4mm}
    \caption{The proposed CharNet can directly output bounding boxes of words and characters, with corresponding character labels in one pass.}
    \label{fig:prediction_results}
\end{figure}

Text reading in natural images has long been considered as two separate tasks: text detection and recognition, which are
implemented sequentially. The two tasks have been advanced individually by the success of deep neural networks. Text detection aims to predict a bounding box for each text instance (e.g., typically a word) in natural images, and current leading approaches are mainly extended from object detection or segmentation frameworks, such as \cite{lyu2018mask, zhou2017east, liu2018fots}. Built on text detection, the goal of text recognition is to recognize a sequence of character labels from an cropped image patch including a text instance. Generally, it can be cast into a sequence labeling problem, where various recurrent models with CNN-extracted features have been developed, with state-of-the-art performance achieved \cite{shi2018aster, cheng2017iccv, shi2017end, he2016reading}.

However, the two-step pipeline often suffers from a number of limitations. First, learning the two tasks independently would result in a sub-optimization problem, making it difficult to fully explore the potential of text nature. For example, text detection and recognition can work collaboratively by providing strong context and complementary information to each other, which is critical to improving the performance, as substantiated by recent work \cite{he2018end, liu2018fots}. Second, it often requires to implement multiple sequential steps, resulting in a relatively complicated system, where the performance of text recognition is heavily relied on text detection results.

Recent effort has been devoted to developing a unified framework that implements text detection and recognition simultaneously \cite{he2018end, liu2018fots, lyu2018mask}. For example, in \cite{he2018end} and \cite{liu2018fots},  text detection models were extended to joint detection and recognition, by adding a new RNN-based branch for recognition, leading to the state-of-the-art performance on end-to-end (E2E) text recognition. These approaches can achieve joint detection and recognition using a single model, but they are in the family of two-stage framework and thus have the following limitations.
Firstly, the recognition branch often explores a RNN-based sequential model, which is difficult to optimize jointly with the detection task, by requiring a significantly larger amount of training samples. Thus the performance is heavily depended on a well-designed but complicated training scheme (e.g., \cite{he2018end} and \cite{li2017towards}). This is the central issue that impedes the development of a united framework.
Secondly, current two-stage framework commonly involves RoI cropping and pooling, making it difficult to crop an accurate text region for feature pooling, where a large amount of background information may be included. This inevitably leads to significant performance degradation on recognition task, particularly for multi-orientation or curved text.


%

To overcome the limitations of RoI cropping and pooling for two-stage framework, He \emph{et al.} \cite{he2018end} proposed a text-alignment layer to precisely compute the convolutional
features for a text instance of arbitrary orientation, which boosted the performance. In \cite{liu2018fots}, multiple affinity transformations were applied to the convolutional features for enhancing text information in the RoI regions. However, these methods failed to work on curved text.
In addition, many high-performance models consider words (for English) as detection units,
but word-level detection often requires to cast text recognition into a sequence labelling
 problem, where a RNN model with additional modules, such as CTC \cite{graves2006connectionist,he2016aaai,shi2017pami} or attention mechanism \cite{shi2018aster, cheng2017iccv,bai2018cvpr,he2018end}, was applied. Unlike English, words are not clearly distinguishable in some languages such as Chinese, where text instances can be defined and separated more clearly by characters.
Therefore, characters are more clearly-defined elements that generalize better over various languages. Importantly, character recognition is straightforward, and can be implemented with a simple CNN model, rather than using a RNN-based sequential model.

\textbf{Contributions.} In this work, we present Convolotional Character Networks (referred as CharNet) for joint text detection and recognition, by leveraging character as basic unit. Moreover, for the first time, we provide an one-stage CNN model for the joint tasks, with significant performance improvements over the state-of-the-art results achieved by a more complex two-stage framework, such as \cite{he2018end}, \cite{lyu2018mask} and \cite{liu2018fots}. The proposed CharNet implements direct character detection and recognition, jointly with text instance (e.g., word) detection. This allows it to avoid the RNN-based word recognition, resulting in a simple, compact, yet powerful model that directly outputs bounding boxes of words and characters, as well as the corresponding character labels, as shown in Fig.\ref{fig:prediction_results}. Our main contributions are summarized as follows.

Firstly, we propose an one-stage CharNet for joint text detection and recognition, where a new branch for direct character detection and recognition is introduced, and can be integrated seamlessly into existing text detection framework. We explore character as basic unit, which allows us to overcome the main limitations of current two-stage framework using RoI pooling with RNN-based recognition.

Secondly, we develop an iterative character detection method which allows CharNet to transform the character detection capability learned from synthetic data to real-world images. This makes it possible for training CharNet on real-world images, without providing additional char-level bounding boxes.

Thirdly, CharNet consistently outperforms recent two-stage approaches such as \cite{he2018end, lyu2018mask, liu2018fots, sun2018textnet} by a large margin, with improvements of 65.33\%$\rightarrow$71.08\% (generic lexicon) on ICDAR 2015, and 54.0\%$\rightarrow$69.23\% (E2E) on Total-Text. Particularly, it can achieve comparable results, e.g., 67.24\% on ICDAR 2015, even by completely removing a lexicon.
\section{Related Work}
Traditional approaches often regard text detection and recognition as two separate tasks that process sequentially \cite{huang2013,huang2014,tian2016detecting,zhou2017east,he2016reading,shi2017pami}. Recent progress has been made on developing a unified framework for joint text detection and recognition \cite{he2018end,liu2018fots,lyu2018mask}. We briefly review the related studies on text detection, recognition and join of two tasks.

\textbf{Text detection.} Recent approaches for text detection were mainly built on general object detectors with various text-specific modifications. For instance, by building on Region Proposal Networks \cite{ren2015faster}, Tian \etal \cite{tian2016detecting} proposed a Connectionist Text Proposal Network (CTPN) to explore the sequence nature of text, and detect a text instance in a sequence of fine-scale text proposals. Similarly, Shi \etal \cite{shi2017detecting}  developed a method with linking segment which also localizes a text instance in a sequence, with the capability for detecting multi-oriented text. In \cite{zhou2017east}, EAST was introduced by exploring IOU loss \cite{yu2016unitbox} to detect multi-oriented text instances (e.g., words), with impressive results achieved. Recently, a single-shot text detector (SSTD) \cite{he2017single} was proposed by extending SSD object detector \cite{liu2016ssd} to text detection. SSTD encodes text regional attention into convolutional features to enhance text information.


\begin{figure*}[t]
\begin{center}
   \includegraphics[width=0.9\linewidth]{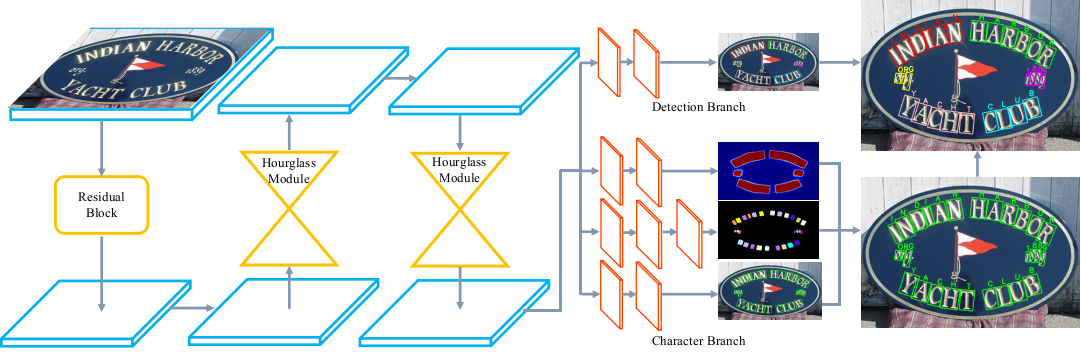}
\end{center}
\vspace{-4mm}
   \caption{Overview of the proposed CharNet, which contains two branches working in parallel: a character branch for direct character detection and recognition, and a detection branch for text instance detection.}
   \label{fig:framework_overview}
\end{figure*}

\textbf{Text recognition.} Inspired from speech recognition, recent work on text recognition commonly cast it into a sequence-to-sequence recognition problem, where recurrent neural networks (RNNs) were employed. For example, He \etal \cite{he2016reading} exploited convolution neural networks (CNNs) to encode a raw input image into a sequence of deep features, and then a RNN is applied to the sequential features for decoding and yielding confidence maps, where connectionist temporal classification CTC \cite{graves2006connectionist} is applied to generate final results. Shi \etal \cite{shi2017end} improved such CNN+RNN+CTC framework by making it end-to-end trainable, with significant performance gain obtained. Recently, the framework was further improved by introducing various attention mechanisms, which are able to encode more character information explicitly or implicitly \cite{shi2018aster, cheng2017iccv,bai2018cvpr,he2018end}.


\textbf{End-to-end (E2E) text recognition.} Recent work attempted to integrate text detection and recognition into a unified framework for E2E text recognition. Li \etal \cite{li2017towards} drew inspiration from Faster R-CNN \cite{ren2015faster} and employed RoI pooling to obtain text features from a detection framework for further recognition.
In \cite{he2018end}, He \etal proposed an E2E framework by introducing a new text-alignment layer with character attention mechanism, leading to significant performance improvements by jointly training two tasks. Similar framework has been developed by Liu \etal in \cite{liu2018fots}. Both works have achieved strong performance on E2E text recognition, but they were built on two-stage models implementing ROI cropping and pooling operations, which may reduce the performance, particularly on the recognition task for multi-orientation or curved text.

Our work is related to character-based approaches for text detection or recognition. Hu \etal proposed a WordSup able to detect text instances at the character level \cite{hu2017wordsup}, while Liu \etal \cite{liu2018char} developed a character-aware neural network for distorted scene text recognition. However, they did not provide a full solution for E2E text recognition. The most closely related work is that of  Mask TextSpotter \cite{lyu2018mask} which is a two-stage character-based framework for E2E recognition, built on recent Mask R-CNN \cite{he2017single}. However, our CharNet has a number of clear distinctions: (1) CharNet is the first one-stage model for E2E text recognition, which is different from the two-stage Mask TextSpotter, where RoI cropping and pooling operations are required; (2) CharNet has a character branch that directly outputs accurate char-level bounding boxes. This enables it to automatically identify characters, allowing it to work in a weakly-supervised manner by using the proposed iterative character detection;
(3) This results in a distinct capability for training CharNet without additional char-level bounding boxes in real-world images, while Mask TextSpotter requires full char-level annotations which are often highly expensive; (4) CharNet achieved consistent and significant performance improvements over Mask TextSpotter, as shown in Table~\ref{tab:results_icdar15} and~\ref{tab:results_totaltext}.

\section{Convolutional Character Networks}
In this section, we describe the proposed CharNet in details.
Then an iterative character detection method is introduced for automatically identifying characters with bounding boxes from real-world images, by leveraging synthetic data. In this work, we use ``text instance" as a higher level concept for text, which can be a word or a text-line, with multi-orientation or curved shape.

%

\subsection{Overview}
As discussed, existing approaches for E2E text recognition are commonly limited by using RoI cropping and pooling, with a RNN-based sequential model for word recognition. The proposed CharNet is an one-stage convolutional architecture consisting of two branches: (1) a character branch designed for direct character detection and recognition, and (2) a text detection branch predicting a bounding box for each text instance in an image.
%
%
The two branches are implemented in parallel, which form an one-stage model for joint text detection and recognition, as shown in Fig. \ref{fig:framework_overview}.
%
%
%
The character branch can be integrated seamlessly into an one-stage text detection framework, resulting in an end-to-end trainable model. Training the model requires both instance-level and char-level bounding boxes with character labels as supervised information. In inference, CharNet can directly output both instance-level and char-level bounding boxes with corresponding character labels in one pass.

Many existing text databases often do not include char-level annotations which are highly expensive to obtain. We develop an iterative learning approach for automatic character detection, which allows us to learn a character detector from synthetic data where full char-level annotations can be generated unlimitedly. Then the learned character detection capability can be transformed and adapted gradually to real-word images. This enables the model with ability to automatically identify characters in real-world images, providing a weakly-supervision learning manner for CharNet.


\textbf{Backbone networks.}
We employ ResNet-50 \cite{he2016deep} and Hourglass \cite{law2018cornernet} networks as backbone for our CharNet framework. For ResNet-50, we follow \cite{zhou2017east}, and make use of the convolutional feature maps with $4\times$ down-sampling ratio as the final convolutional maps to implement text detection and recognition. This results in high-resolution feature maps that enable CharNet to identify extremely small-scale text instances. For Hourglass networks, we stack two hourglass modules, as shown in Fig.~\ref{fig:framework_overview}, and the final feature maps are up-sampled to $\frac{1}{4}$ resolution of the input image.  In this work, we use two variants of Hourglass networks, Hourglass-88 and Hourglass-57. Hourglass-88 is modified from Hourglass-104 in \cite{law2018cornernet} by removing two down-sampling stages and reducing the number of layers in the last stage of each hourglass module by half. Hourglass-57 is constructed by further removing half number of layers in each stage of hourglass modules. Notice that, for both variants, we do not employ the intermediate supervision as did in CornerNet \cite{law2018cornernet}.

\subsection{Character Branch}
%
Existing RNN-based recognition methods were commonly built on word-level optimization with a sequential model, which has a significantly larger search space than direct character classification. This inevitably makes the models more complicated and difficult to train by requiring a significantly longer training time with a large amount of training samples.
%
Recent work, such as \cite{shi2018aster, cheng2017iccv, he2018end}, had shown that the performance of RNN-based methods can be improved considerably by introducing char-level attention mechanism which is able to encode strong character information implicitly or explicitly. This enables the models to have the ability to identify characters more accurately,
and essentially adds additional constraints to the models which in turn reduce the search space, leading to performance boost.
\emph{This suggests that precise identification of characters is of great importance to RNN-based text recognition, which inspired the current work to simplify it into direct character recognition with an automatic character localization mechanism, resulting in a simple yet powerful one-stage fully convolutional model for E2E text recognition.}


To this end, we introduce a new character branch which has the functions of direct character detection and recognition. The character branch uses character as basic unit for detection and recognition, and outputs char-level bounding boxes as well as the corresponding character labels. Specifically, the character branch is a stack of convolutional layers, which move densely over the final feature maps of the backbone.
It has the input features maps with $\frac{1}{4}$ spatial resolution of the input image. This branch contains three sub-branches, for text instance segmentation, character detection and character recognition, respectively. The text instance segmentation sub-branch and character detection sub-branch have three convolutional layers with filter sizes of $3\times3$, $3\times3$ and $1\times1$, respectively. The character recognition sub-branch has four convolutional layers with one more $3\times3$ convolutional layer.

Text instance segmentation sub-branch exploits a binary mask as supervision, and outputs 2-channel feature maps indicating text or non-text probability at each spatial location. Character detection sub-branch outputs 5-channel feature maps, estimating a character bounding box at each spatial location. By following EAST \cite{zhou2017east}, each character bounding box is parameterized by five parameters, indicating the distances of current location to the top, bottom, left and right sides of the bounding box, as well as the orientation of bounding-box. In character recognition sub-branch, character labels are predicted densely over the input feature maps, generating 68-channel probability maps. Each channel is a probability map for a specific character class among 68 character classes, including 26 English characters, 10 digital numbers and 32 special symbols. All of the output feature maps from three sub-branches have the same spatial resolution, which is exactly the same as that of the input feature maps ($\frac{1}{4}$ of the input image).
Finally, the char-level bounding boxes are generated by keeping the bounding boxes having a confident value over 0.95.  Each generated bounding box has a corresponding character label, which is computed at the corresponding spatial location from the 68-channel classification maps - by using the maximum of the computed softmax scores.



Training character branch requires char-level bounding boxes with the corresponding character labels. Compared to word-level annotations, acquiring char-level labels with bounding boxes is much more expensive and would significantly increase labor cost. To avoid such additional cost, we develop an iterative character detection mechanism which is described in Section~\ref{sec:icd}.


\subsection{Text Detection Branch}
Text detection branch is designed to identify text instances at a higher level concept, such as words or text-lines. It provides strong context information which is used to group the detected characters into text instances. Because directly grouping characters by using characters information (e.g., character locations or geometric features) is heuristic and complicated when multiple text instances are located closely within a region, particularly for text instances with multiple orientations or in a curved shape.
Our text detection branch can be defined in different forms subjected to the type of text instances, and existing instance-level text detectors can be adapted with minimum modification. We take text detectors for multi-orientation words or curved text-lines as examples.

\textbf{Multi-Orientation Text.} We simply modify EAST detector \cite{zhou2017east} as our text detection branch, which contains two sub-branches for text instance segmentation and instance-level bounding box regression using IoU loss. The predicted bounding boxes are parameterized by five parameters including 4 scalars for a bounding box with an orientation angle. We compute dense prediction at each spatial location of the feature maps by using two $3\times3$ convolutional layers, followed by another $1\times1$ convolutional layer. Finally, the text detection branch outputs 2-channel feature maps indicating text or non-text probability, and 5-channel detection maps for bounding boxes with orientation angles. We keep the bounding boxes having a confident value over 0.95.

\textbf{Curved Text.}
For curved text, we modify Textfield in \cite{xu2019textfield} by using a direction field, which encodes the direction information that points away from text boundary. The direction field is used to separate adjacent text instances, and can be predicted by a new branch in parallel with text detection branch and character branch. This branch is composed of two $3\times3$ convolutional layers, followed by another $1\times1$ convolutional layer.

\textbf{Generation of Final Results.} The predicted instance-level bounding boxes are applied to group the generated characters into text instances. We make use of a simple rule, by assigning a character to a text instance if the character bounding box have an overlap (e.g., with $>0$ IoU) with an instance-level bounding box. The final outputs of our CharNet are bounding boxes of both text instances and characters, with the corresponding character labels.

\subsection{Iterative Character Detection} \label{sec:icd}
Training our model requires both char-level and word-level bounding boxes as well as the corresponding character labels. However, char-level bounding boxes are expensive to obtain and are not available in many existing benchmark datasets such as ICDAR 2015 \cite{karatzas2015icdar} and Total-Text \cite{ch2017total}. We develop an iterative character detection method that enables our model to have capability for identifying characters by leveraging synthetic data, such as Synth800k \cite{gupta2016synthetic}, where multi-level supervised information can be generated unlimitedly.
This allows us to train CharNet in a weakly-supervised manner by just using instance-level annotations from real-world images.
%
%

A straightforward approach is to train our model directly with synthetic images, and then run inference on real-world images. However, it has a large domain gap between the synthetic images and real ones, and therefore the model trained from synthetic images is difficult to work directly on the real-world ones, as shown in Table \ref{tab:icdar15_only_synth}, where low performance is obtained on both text detection and E2E recognition.
We observed that a text detector has relatively stronger generalization capability than a text recognizer. As shown in \cite{tian2016detecting}, a text detector trained solely on English and Chinese data can work reasonably on other languages, which inspired us to explore the generalization ability of a character detector to bridge the gap between the two domains.

\begin{table}[!t]
	\centering
	\begin{tabular}{l|c|c|c}
		\hline
		Method & w/ Real. & Detection & E2E \\
		\hline
        CharNet R-50 & & 65.38 & 33.69 \\
        CharNet R-50 & \checkmark & 89.70 & 62.18 \\
        CharNet H-57 & & 65.19 & 39.43 \\
        CharNet H-57 & \checkmark & 89.66 & 66.92 \\
        CharNet H-88 & & 65.11 & 39.94 \\
        CharNet H-88 & \checkmark & 90.97 & 69.14 \\
		\hline
	\end{tabular}
	\vspace{-2mm}
	\caption{Performance of CharNet with various backbone networks on ICDAR 2015. ``Real." denotes ``CharNet trained on real-world images with the proposed iterative character detection". Detection is compared by using F-measure.}
	\label{tab:icdar15_only_synth}
\end{table}

\begin{figure}[!t]
    \begin{center}
        \includegraphics[height=4.3cm]{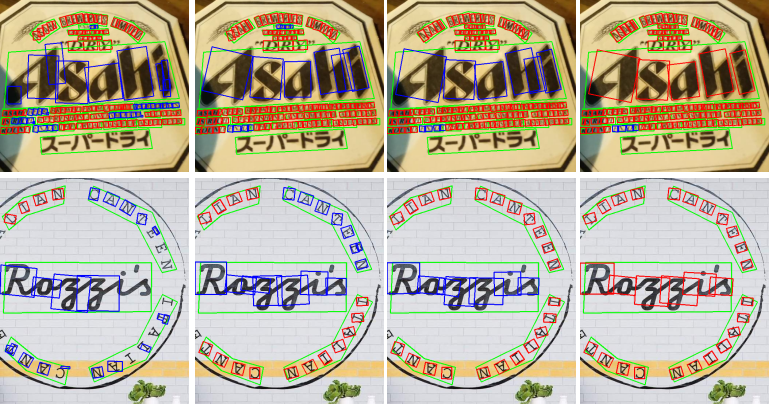}
    \end{center}
    \vspace{-4mm}
    \caption{Character bounding boxes generated at 4 interactive steps from left to right.  Red boxes indicate the identified ``correct" ones by our rule, while blue boxes mean invalid ones, which are not collected for training in next step.}
    \label{fig:iteration}
\end{figure}

 Our intuition is to gradually improve the generalization capability of model which is initially trained from synthetic images where full char-level annotations are provided, and the key is to transform the capability of \emph{character detection} learned from the synthetic data to real-world images.  We develop an iterative process by gradually identifying the ``correct" char-level bounding boxes from real-world images by the model itself. We make use of a simple rule that identifies a group of char-level bounding boxes as ``correct" if \emph{the number of character bounding boxes in a text instance is exactly equal to the number of character labels in the provided instance-level transcript}. Note that instance-level transcripts (e.g., words) are often provided in existing datasets for E2E text recognition.
%
%
%
The proposed iterative character detection are described as follows.
\begin{itemize}
\item[--] (i) We first train an initial model on synthetic data, where both char-level and instance-level annotations are available to our CharNet. Then we apply the trained model to the training images from a real-world dataset, where char-level bounding boxes are predicted by the learned model.
\item[--] (ii) We explore the aforementioned rule to collect the ``correct" char-level bounding boxes detected in real-world images, which are used to further train the model with the corresponding transcripts provided. Note that we do not use the predicted character labels, which are not fully correct and would reduce the performance in our experiments.


\item[--] (iii) This process is implemented iteratively to enhance the model capability gradually for character detection, which in turn continuously improves the quality of the identified characters, with an increasing number of the ``correct" char-level bounding boxes generated, as shown in Fig. \ref{fig:iteration} and Table \ref{tab:iterative_annotation_generation}.
\end{itemize}



\section{Experiments, Results and Comparisons}
Our CharNet is evaluated on  three standard benchmarks: ICDAR 2015 \cite{karatzas2015icdar}, Total-Text \cite{ch2017total}, and ICDAR MLT 2017 \cite{nayef2017icdar2017}.
\textbf{ICDAR 2015} includes 1,500 images collected by using Google Glasses. The training set has 1,000 images, and the remaining 500 images are used for evaluation. This dataset is challenging due to the presence of multi-orientated and very small-scale text instances.
\textbf{Total-Text} consists of 1,555 images with a variety of text types including horizontal, multi-oriented, and curved text instances. The training split and testing split have 1,255 images and 300 images, respectively.
\textbf{ICDAR MLT 2017} is a large-scale multi-lingual text dataset, which contains 7,200 training images, 1,800 validation images, and 9,000 testing images. 9 languages are included in total.

\subsection{Implementation Details}
Similar to recent work in \cite{he2018end, liu2018fots}, our CharNet is trained on both synthetic data and real-world data. The proposed iterative character detection is implemented by using 4 iterative steps. At the first step, CharNet is trained on synthetic data, Synth800k \cite{gupta2016synthetic}, for 5 epochs, where both char-level and word-level annotations are available. We use a mini-batch of 32 images, with 4 images per GPU. On the synthetic data, we set a base learning rate of $0.0002$, which is reduced according to $lr_{base} \times(1 - \frac{iter}{max\_iter})^{power}$ with $power=0.9$, by following \cite{chen2018deeplab}. The remained three iterative steps are implemented on real-world data, by training CharNet for 100, 400 and 800 epochs respectively, on the training set of a benchmark provided, e.g., ICDAR 2015 \cite{karatzas2015icdar} or Total-Text \cite{ch2017total}. On the real-world data, we set a base learning rate of $0.002$, and use the char-level bounding boxes generated by the model trained from the previous step. We make use of similar data augmentation as \cite{liu2018fots} and OHEM \cite{shrivastava2016training}.

\subsection{On Iterative Character Detection}
\begin{table}[!t]
	\centering
	\begin{tabular}{c|c|c|c|c}
	    \hline
		Step & \# Words & Ratio (\%) & E2E & \# Epochs\\
		\hline
		0 & 6033 & 64.95 & 39.3 & 5 \\
		1 & 8262 & 88.94 & 62.9 & 100\\
        2 & 8494 & 91.44 & 65.0 & 400\\
        3 & \textbf{8606} & \textbf{92.65} & \textbf{66.1} & 800\\
        \hline
	\end{tabular}
	\vspace{-2mm}
	\caption{4-step iterative character detection with CharNet. ``\# Words" is the number of words identified as ``correct" at each step iterative learning. ``Ratio" denotes the ratio of the ``correct" words to all words in the training images from Total-Text. ``\# Epochs" indicates the number of training epochs for each iterative step. At the Step 0, CharNet is trained on synthetic data for 5 epochs, while Step 1-3 are implemented on real-world images.  ``E2E" means ``End-to-End Recognition with F-measure".}
	\label{tab:iterative_annotation_generation}
\end{table}

Interactive character detection is an important function for CharNet that allows us to train the model on real-world images by only using text instance-level annotations. Thus accurate identification of characters is critical to the performance of CharNet. We evaluate the iterative character detection with CharNet by using various backbone networks on ICDAR 2015. Results are reported in Table~\ref{tab:icdar15_only_synth}.
As can be found, CharNet has low performance on both text detection and E2E recognition when we directly apply the model trained from synthetic data to testing images from ICDAR 2015, due to a large domain gap between the two data sets.
The performance can be improved considerably by training CharNet on real-world data with iterative character detection, which demonstrates its efficiency.



We further investigate the capability of our model for identifying the ``correct" characters in real-world images. Experiments were conducted on Total-Text. In this experiment, the ``correct" characters are grouped into words, and we calculate the number of correctly-detected words at each iterative step. As shown in Table~\ref{tab:iterative_annotation_generation}, at the step 0, when CharNet is only trained on synthetic data, only $64.95\%$ words are identified as ``correct" from real-world training images.
%
%
Interestingly, this number increases immediately from $64.95\%$ to $88.94\%$ at the step 1, when the proposed iterative character detection is applied. This also leads to a significant performance improvement, from 39.3\% to 62.9\% on E2E text recognition. The iterative training continues
until the number of the identified words dose not increase further. Finally, our method is able to collect $92.65\%$ correct words from real-world images by implementing 4 iterative steps in total. We argue that this number of char-level annotations learned automatically by model is enough to train our CharNet, as evidenced by the state-of-the-art performance obtained, which is shown next.



\subsection{Results on Text Detection}
\begin{figure}[!t]
    \begin{center}
        \includegraphics[height=3.2cm]{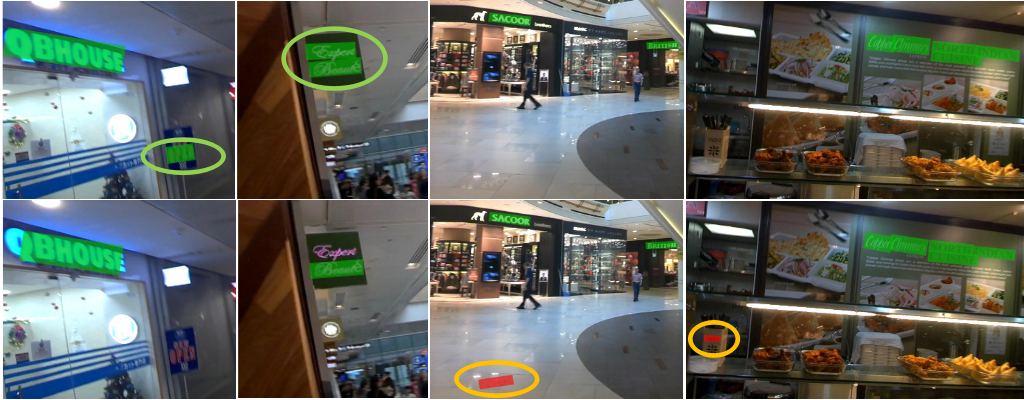}
    \end{center}
    \vspace{-4mm}
    \caption{CharNet improves both recall and precision on text detection by jointly learning with  character recognition.}
    \label{fig:recog_help_det}
\end{figure}

We evaluate the performance of CharNet on text detection task.
%
%
To make a fair comparison, we use the same backbone ResNet-50 as FOTS \cite{liu2018fots}. As shown in Table~\ref{tab:icdar15_det_results_vs_fots}, our CharNet achieves comparable performance with FOTS when both methods are trained without recognition branch. By jointly optimizing the model with text recognition, CharNet improves the detection performance by $4.13\%$, from a F-measure of 85.57\% to 89.70\%, which is more significant than $2.68\%$ performance gain achieved by FOTS. It suggests that our one-stage model allows text detection and recognition to work more effectively and collaboratively. This enables CharNet with higher capability for identifying extremely challenging text instances with stronger robustness which also reduces false detections, as shown in Fig. \ref{fig:recog_help_det}. In addition, CharNet also has a performance improvement of  $87.00\%\rightarrow89.70\%$ on F-measure over that of \cite{he2018end} which uses a PVAnet \cite{hong2016} as backbone with multi-scale implementation.

\begin{figure}[!t]
    \begin{center}
        \includegraphics[width=8.5cm]{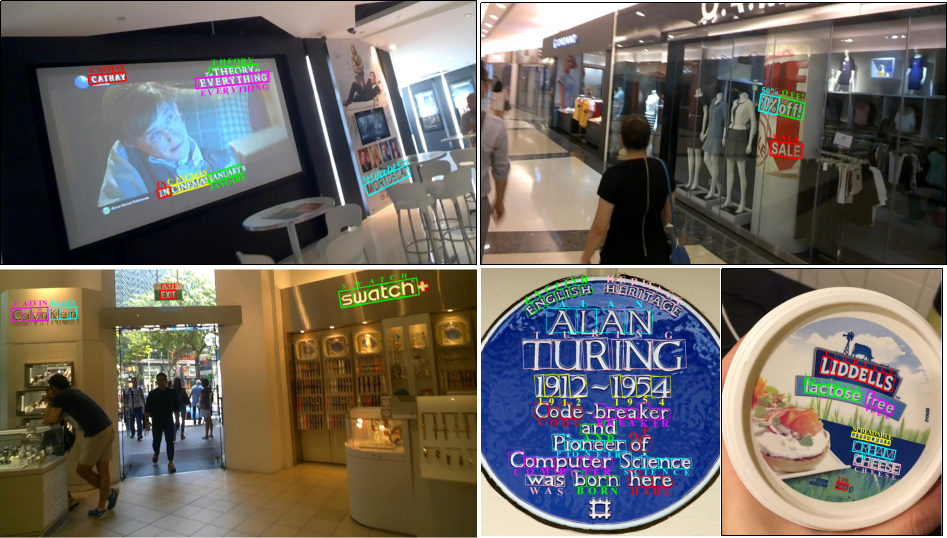}
    \end{center}
    \vspace{-5mm}
    \caption{Full results by CharNet.}
    \label{fig:show_more_result}
\end{figure}

Moreover, our one-stage CharNet achieves new stage-of-the-art performance on text detection on all three benchmarks, which improves recent strong baseline (e.g., He \emph{et al.} \cite{he2018end}, FOTS \cite{liu2018fots} and TextFiled \cite{xu2019textfield}) by a large margin. For example, on single-scale case, the improvements on F-measure are: $87.99\%\rightarrow90.97\%$ on ICDAR 2015 (in Table~\ref{tab:results_icdar15}), $80.3\%\rightarrow85.6\%$ on the Total-Text for curved text (in Table~\ref{tab:results_totaltext}), and 67.25\%$\rightarrow$75.77\% on ICDAR 2017 MLT (in Table~\ref{tab:mlt17_results}). Notice that CharNet is designed by using characters as basic unit. This natural property allows it to be easily adapted to curved text, where FOTS is difficult to work reliably. TextFiled was designed specifically for curved text but only has a F-measure of 82.4\% on ICDAR 2015. Several examples for detecting challenging text instances are presented in Fig.~\ref{fig:show_more_result}.

\begin{table}[!t]
	\centering
	\begin{tabular}{l|c|c|c|c|c}
		\hline
		Method & Rec. & R & P & F & Gain \\
		\hline
        He \emph{et al.} \cite{he2018end}& & 83.00 & 84.00 & 83.00 & - \\
        He \emph{et al.} \cite{he2018end} & \checkmark & 86.00 & 87.00 & 87.00 & +4.00 \\
        FOTS \cite{liu2018fots} & & 82.04 & 88.84 & 85.31 & - \\
        FOTS \cite{liu2018fots} & \checkmark & 85.17 & 91.00 & 87.99 & +2.68 \\
        CharNet & & 81.37 & 90.23 & 85.57 & - \\
        CharNet & \checkmark & \textbf{88.30} & \textbf{91.15} & \textbf{89.70} & \textbf{+4.13} \\
        \hline
	\end{tabular}
	\vspace{-2mm}
	\caption{Detection performance on ICDAR 2015. ResNet-50 was used by both FOTS and CharNet as backbone, while PVAnet \cite{hong2016} was applied in \cite{he2018end}. ``Rec." denotes ``Recognition". ``Gain" is the performance gain obtained by joint optimization with text recognition.  ``R", ``P", ``F" indicate ``Recall", ``Precision", ``F-measure".}
	\label{tab:icdar15_det_results_vs_fots}
\end{table}

\begin{table*}[!t]
    \centering
	\begin{tabular}{l|c|c|c|c|l|c|c|c|c}
	    \hline
		\multirow{2}{*}{Method} & \multirow{2}{*}{Params} & \multicolumn{3}{c|}{Detection} & \multirow{2}{*}{Method} & \multicolumn{4}{c}{End-to-End Recognition} \\
		\cline{3-5} \cline{7-10} & & R & P & F & & S & W & G & N \\
		\hline
		\multicolumn{10}{c}{Single Scale}\\
 		\hline
 		WordSup \cite{hu2017wordsup} & - & 77.03 & 79.33 & 78.16 & Neumann \etal \cite{neumann2016real} & 35.00 & 20.00 & 16.00 & - \\
  		EAST \cite{zhou2017east} & - & 78.33 & 83.27 & 80.72 & Deep text spotter \cite{busta2017deep} & 54.00 & 51.00 & 47.00 & - \\
		R2CNN \cite{jiang2017r2cnn} & - & 79.68 & 85.62 & 82.54 & TextProp.+DictNet \cite{hochreiter1997long, zhang2015symmetry} & 53.30 & 49.61 & 47.18 & - \\
		Mask TextSpotter \cite{lyu2018mask} * & - & 81.00 & 91.60 & 86.00 & Mask TextSpotter \cite{lyu2018mask} * & 79.30 & 73.00 & 62.40 & - \\
		FOTS R-50 \cite{liu2018fots} & 34.98 M & 85.17 & 91.00 & 87.99 & FOTS R-50 \cite{liu2018fots} & 81.09 & 75.90 & 60.80 & - \\
        \hline
		CharNet R-50 & 26.48 M & 88.30 & 91.15 & 89.70 & CharNet R-50 & 80.14 & 74.45 & 62.18 & 60.72 \\
		CharNet H-57 & 34.96 M & 88.88 & 90.45 & 89.66 & CharNet H-57 & 81.43 & 77.62 & 66.92 & 62.79 \\
		CharNet H-88 & 89.21 M & \textbf{89.99} & \textbf{91.98} & \textbf{90.97} & CharNet H-88 & \textbf{83.10} & \textbf{79.15} & \textbf{69.14} & \textbf{65.73} \\
		\hline \multicolumn{10}{c}{Multi-Scale} \\ \hline
		He \etal \textit{MS} \cite{he2018end} & - & 86.00 & 87.00 & 87.00 & He \etal \textit{MS} \cite{he2018end} & 82.00 & 77.00 & 63.00 & - \\
		FOTS R-50 \textit{MS} \cite{liu2018fots} & 34.98 M & 87.92 & 91.85 & 89.84 & FOTS R-50 \textit{MS} \cite{liu2018fots} & 83.55 & 79.11 & 65.33 & - \\
		\hline
		CharNet R-50 \textit{MS} & 26.48 M & 90.90 & 89.44 & 90.16 & CharNet R-50 \textit{MS} & 82.46 & 78.86 & 67.64 & 62.71 \\
		CharNet H-57 \textit{MS} & 34.96 M & \textbf{91.43} & 88.74 & 90.06 & CharNet H-57  \textit{MS} & 84.07 & 80.10 & 69.21 & 65.26 \\
		CharNet H-88 \textit{MS} & 89.21 M & 90.47 & \textbf{92.65} & \textbf{91.55} & CharNet H-88  \textit{MS} & \textbf{85.05} & \textbf{81.25} & \textbf{71.08} & \textbf{67.24} \\
		\hline
	\end{tabular}
	\vspace{-2mm}
	\caption{Results on ICDAR 2015.  ``R-*" and ``H-*" denote ``ResNet-*" and ``Hourglass-*".  ``\textit{MS}" means multi-scale inference.  ``R", ``P", ``R" are ``Recall", ``Precision", ``F-measure". ``S", ``W", ``G" and ``N" mean F-measure using ``Strong", ``Week", ``Generic" and ``None" lexicon. }
	\label{tab:results_icdar15}
\end{table*}

\begin{table}[!t]
    \centering
	\begin{tabular}{l|c|c|c|c}
	    \hline
		\multirow{2}{*}{Method} & \multicolumn{3}{c|}{Detection} & \multirow{2}{*}{E2E} \\
		\cline{2-4} & R & P & F \\
        \hline
        Textboxes \cite{liao2017textboxes} & 45.5 & 62.1 & 52.5 & 36.3 \\
  		Mask TextSpotter \cite{lyu2018mask} & 55.0 & 69.0 & 61.3 & 52.9 \\
  		TextNet \cite{sun2018textnet} & 59.5 & 68.2 & 63.5 & 54.0 \\
  		TextFiled \cite{xu2019textfield} & 79.9 & 81.2 & 80.6 & - \\
        \hline\hline
		CharNet H-57 & 81.0 & 88.6 & 84.6 & 63.6 \\
		CharNet H-88 & 81.7 & \textbf{89.9} & 85.6 & 66.6 \\
		CharNet H-57 \textit{MS} & \textbf{85.0} & 87.3 & 86.1 & 66.2 \\
		CharNet H-88 \textit{MS} & \textbf{85.0} & 88.0 & \textbf{86.5} & \textbf{69.2} \\
		\hline
	\end{tabular}
	\vspace{-2mm}
	\caption{Results on Total-Text.  ``H-*" denotes ``Hourglass-*". ``\textit{MS}" indicates multi-scale inference.  ``R", ``P", ``R" are ``Recall", ``Precision", ``F-measure".  ``E2E" is ``End-to-End Recognition using F-measure".}
	\label{tab:results_totaltext}
\end{table}

\begin{table}[!h]
	\centering
	\begin{tabular}{l|c|c|c}
		\hline
		Method & R & P & F \\
		\hline
        SARI\_FDU\_RRPN \cite{ma2018arbitrary} & 55.50 & 71.17 & 62.37 \\
        SCUT\_DLVClab & 54.54 & 80.28 & 64.96 \\
        FOTS \cite{liu2018fots} & 57.51 & 80.95 & 67.25 \\
        FOTS \textit{MS} \cite{liu2018fots} & 62.30 & \textbf{81.86} & 70.75 \\
		\hline\hline
		CharNet R-50 & 70.10 & 77.07 & 73.42 \\
		CharNet H-88 & \textbf{70.97} & 81.27 & \textbf{75.77} \\
		\hline
	\end{tabular}
	\vspace{-2mm}
	\caption{Text detection on ICDAR 2017 MLT. ``R-*" and ``H-*" denote ``ResNet-*" and ``Hourglass-*".  ``R", ``P" and ``F" represent ``Recall", ``Precision" and ``F-measure". ``\textit{MS}" indicates multi-scale inference.}
	\label{tab:mlt17_results}
\end{table}



\subsection{Results on End-to-End Text Recognition}
For E2E text recognition task, we compare our CharNet with recent state-of-the-art methods on ICDAR 2015 \cite{karatzas2015icdar} and Total-Text \cite{ch2017total}.

\paragraph{ICDAR 2015.} As shown in Table~\ref{tab:results_icdar15}, by using a same backbone ResNet-50, our CharNet has comparable results with Mask TextSpotter \cite{lyu2018mask}. However, Mask TextSpotter has significant performance improvements by using additional char-level manual annotations on real-world images, with a weighted edit distance applied to a lexicon, e.g., $76.1\% \rightarrow 79.3\%$ (S), $67.1\% \rightarrow 73.0\% $ (W) and $56.7\% \rightarrow 62.4\%$ (G) on E2E recognition. Furthermore, CharNet also outperforms FOTS by $1.38\%$ in terms of generic lexicon. Unlike FOTS, which makes use of a heavy recognition branch with 6.31M parameters, our one-stage model only employs a light-weight CNN-based character branch with 1.19M parameters. Importantly, our model can work reliably without a lexicon, with performance of $60.72\%$, which is comparable to $60.72\%$ of FOTS with a generic lexicon. These lexicon-free results demonstrate the strong capability of our CharNet, making it better applicable to real-world applications where a lexicon is not always available.


We further employ Hourglass-57 \cite{law2018cornernet} as backbone, which has the similar number of model parameters compared to FOTS (34.96M v.s. 34.98M). As shown in Table \ref{tab:results_icdar15}, our CharNet outperforms FOTS by $6.12\%$ with generic lexicon. With a more powerful Hourglass-88, we set a new state-of-the-art single-scale performance on the benchmark, and improve both Mask TextSpotter and FOTS considerably in all terms. Finally, with multi-scale inference, CharNet surpasses the previous best results \cite{liu2018fots} by a large margin, e.g., from 65.33\% to 71.08\% with generic lexicon.

\paragraph{Total-Text.} We conduct experiments on Total-text to show that the capability of our CharNet on curved text. We employ the protocol described in \cite{ch2017total} to evaluate the performance of text detection, and follow the evaluation protocol presented in \cite{lyu2018mask} for E2E recognition. No lexicon is used in E2E recognition. As shown in Table~\ref{tab:results_totaltext}, CharNet outperforms current state-of-the-art methods by $5.9\%$ F-measure on text detection, and $15.2\%$ on E2E recognition.
Compared to character-based method, Mask TextSpotter \cite{lyu2018mask}, our CharNet can obtain even larger performance improvements on curved text.

\section{Conclusions}
We have presented an one-stage CharNet for E2E text recognition. We introduce a new branch for direct character recognition, which can be integrated seamlessly into text detection framework. This results in the first one-stage fully convolutional model that implements two tasks jointly, setting it apart from existing RNN-integrated two-stage framework. We demonstrate that with CharNet, the two tasks can be trained more effectively and collaboratively, leading to significant performance improvements. Furthermore, we develop an iterative character detection able to transfer the character detection capability learned from synthetic data to real-world images.
In addition, CharNet is compact with less parameters, and can work reliably on curved text.
Extensive experiments were conducted on ICDAR 2015, MTL 2017 and Total-text, where CharNet consistently outperforms existing approaches by a large margin.

{
\bibliographystyle{ieee}
\bibliography{egbib}
}
\end{document}